\pdfoutput=1

\documentclass[11pt]{article}

\usepackage[table]{xcolor}
\usepackage[final]{acl}

\usepackage{adjustbox}
\usepackage{times}
\usepackage{latexsym}
\usepackage{amsmath}
\usepackage{booktabs}
\usepackage{multirow}
\usepackage{multicol}
\usepackage{array}
\usepackage{makecell}
\usepackage[shortlabels]{enumitem}
\usepackage[normalem]{ulem}
\useunder{\uline}{\ul}{}
\usepackage{subcaption}
\usepackage{geometry}
\usepackage{graphicx}

\usepackage[T1]{fontenc}

\usepackage[utf8]{inputenc}

\usepackage{microtype}

\usepackage{inconsolata}

\newcommand{\system}{\textsc{Zebra}}
\newcommand{\kb}{\textsc{Zebra-KB}}

\title{ZEBRA: \underline{Z}ero-Shot \underline{E}xample-\underline{B}ased \underline{R}etrieval \underline{A}ugmentation\\for Commonsense Question Answering}

\author{Francesco Maria Molfese\thanks{\ Equal contribution.}, Simone Conia\footnotemark[1], Riccardo Orlando
 \normalfont{and} {\bf Roberto Navigli}\\
Sapienza NLP Group, Sapienza University of Rome\\
\texttt{\{molfese, conia, orlando, navigli\}@diag.uniroma1.it}}

\begin{document}
\maketitle

\begin{abstract}
    Current Large Language Models (LLMs) have shown strong reasoning capabilities in commonsense question answering benchmarks, but the process underlying their success remains largely opaque. As a consequence, recent approaches have equipped LLMs with mechanisms for knowledge retrieval, reasoning and introspection, not only to improve their capabilities but also to enhance the interpretability of their outputs. However, these methods require additional training, hand-crafted templates or human-written explanations. To address these issues, we introduce \system{}, a zero-shot question answering framework that combines retrieval, case-based reasoning and introspection and dispenses with the need for additional training of the LLM. Given an input question, \system{} retrieves relevant question-knowledge pairs from a knowledge base and generates new knowledge by reasoning over the relationships in these pairs. This generated knowledge is then used to answer the input question, improving the model's performance and interpretability. We evaluate our approach across 8 well-established commonsense reasoning benchmarks, demonstrating that \system{} consistently outperforms strong LLMs and previous knowledge integration approaches, achieving an average accuracy improvement of up to 4.5 points.
\end{abstract}

\section{Introduction}\label{sec:intro}


\begin{figure}[t!]
    \centering
    \includegraphics[width=\linewidth]{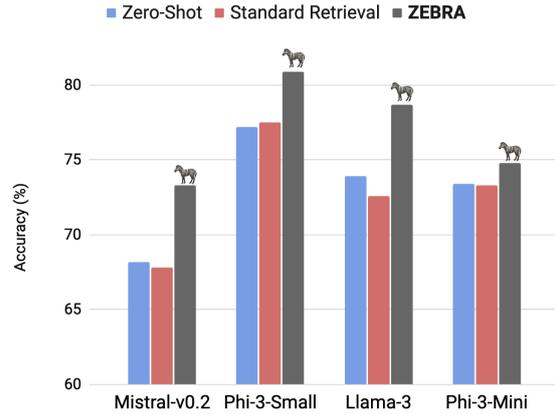}
    \caption{Performance benefits of using \system{} against standard retrieval augmentation methods for commonsense reasoning across four Large Language Models.}
    \label{fig:zebra-chart}
\end{figure}

Over recent years, the research community has explored how to improve the reasoning capabilities of language models and the interpretability of their predictions, with many approaches relying on knowledge augmentation~\cite{liu-etal-2022-generated,zhang2022greaselm,yu-etal-2022-retrieval,liu-etal-2023-crystal}.
For instance, \citet{das-etal-2021-case} introduced \textit{case-based reasoning}~\cite{aamodt1994case} into the training process of a knowledge-based question answering system to guide its reasoning.
At the same time, \citet{shwartz-etal-2020-unsupervised} and \citet{liu-etal-2022-generated} investigated how to prompt LLMs to generate useful knowledge via hand-crafted templates or human-written explanations,
while \citet{yu-etal-2022-retrieval} introduced an approach to teach an LLM to reason over a knowledge base of commonsense knowledge, which is accessed via a retriever.
Furthermore, \citet{liu-etal-2022-rainier} introduced the concept of \textit{knowledge introspection}, which is the process of generating contextually relevant knowledge in response to given questions.

However, some of these approaches rely on commonsense knowledge bases, which are finite by definition and, therefore, may not include the exact information needed to augment the input for downstream tasks.
Moreover, most of the aforementioned approaches require the user to train -- or, at least, fine-tune -- existing LLMs; not only is this process resource-intensive but it also leads to task-specific models, which may result in sub-optimal performance outside the domain of their training datasets.
To address these issues, we introduce \system{}, a zero-shot framework for commonsense reasoning and question answering that aims to combine the benefits of knowledge retrieval, case-based reasoning, and introspection without fine-tuning the underlying LLM.
\system{} stems from two observations: first, direct retrieval of commonsense facts may provide useful hints, but it is limited by the finite nature of knowledge bases and the noise therein; second, introspection can generate contextually relevant knowledge which is tailored to the input question, but this is limited to what the LLM already ``knows''.
In contrast, rather than directly retrieving or generating knowledge for the specific input question, \system{}: i) retrieves one or more \textit{examples} -- also referred to as \textit{cases} in the literature -- that elicit a correct reasoning process for the input question based on commonsense, ii) generates commonsense knowledge tailored for the input question by following the relationship in the question-knowledge pairs contained in the retrieved examples, and iii) uses the generated knowledge to answer the input question.

We can summarize the contributions of this work as follows:
\begin{itemize}
    \item We introduce \system{}, a \underline{z}ero-shot \underline{e}xample-\underline{b}ased \underline{r}etrieval \underline{a}ugmentation framework for commonsense reasoning and question answering that combines the benefits of knowledge retrieval and introspection while dropping the need for additional training of the LLM.
    \item We create \kb{}, a high-quality silver knowledge base for commonsense question answering, where each entry is composed of a question, a list of choices, and a list of explanations based on commonsense reasoning.
    \item We evaluate \system{} and \kb{} across 8 commonsense reasoning benchmarks, demonstrating that \system{} consistently outperforms the baselines, achieving an average accuracy improvement of up to 4.5 points.
\end{itemize}

We believe that \system{} represents a significant step forward for improving the capabilities of LLMs in commonsense question answering tasks, as outlined in Figure \ref{fig:zebra-chart}.
We release our software and data at \url{https://github.com/sapienzanlp/zebra}.

\section{Related Work}\label{sec:rw}
Over the years, the research community has proposed several approaches for the integration of commonsense knowledge into language models.

\paragraph*{Commonsense generation.}
There is a large body of work that has investigated how to pre-train or fine-tune language models to generate commonsense knowledge based on the information available in commonsense knowledge bases and commonsense corpora.
Notably, \citet{bosselut-etal-2019-comet} introduced COMET, a language model trained using explicit knowledge derived from commonsense knowledge bases, such as ConceptNet~\cite{speer2018conceptnet} and ATOMIC~\cite{sap2019atomic}. 
Similarly, \citet{zhou2020pretraining} proposed CALM, a procedure to pre-train language models by integrating generative and contrastive objectives for learning commonsense.
In addition, other approaches, exemplified by Unicorn~\cite{lourie2021unicorn}, fine-tune generative language models to solve a wide-array of commonsense-based tasks in a multitask fashion.
These methods now represent a milestone in the field of commonsense reasoning, as they have shown themselves to be effective in generating structured commonsense knowledge and reasoning over it.
However, these approaches require training and/or fine-tuning on large quantities of data.
Furthermore, since they generate the answers directly, they remain hard to interpret for humans.

\paragraph*{Knowledge integration.}
Other efforts shifted their focus onto improving the reasoning abilities of language models by grounding them in external commonsense knowledge graphs.
Notable works in this direction include KagNet~\cite{lin-etal-2019-kagnet}, GRF~\cite{ji-etal-2020-language}, QA-GNN~\cite{yasunaga-etal-2021-qa} and GreaseLM~\cite{zhang2022greaselm}, among others, which encode commonsense-based knowledge graphs at the input level to guide the reasoning process, while also increasing its interpretability.
However, these approaches not only still require extensive training and fine-tuning, but their performance is also upper-bounded by the quality and completeness of the knowledge graph.

\begin{figure*}[t!]
    \centering
    \includegraphics[width=\textwidth]{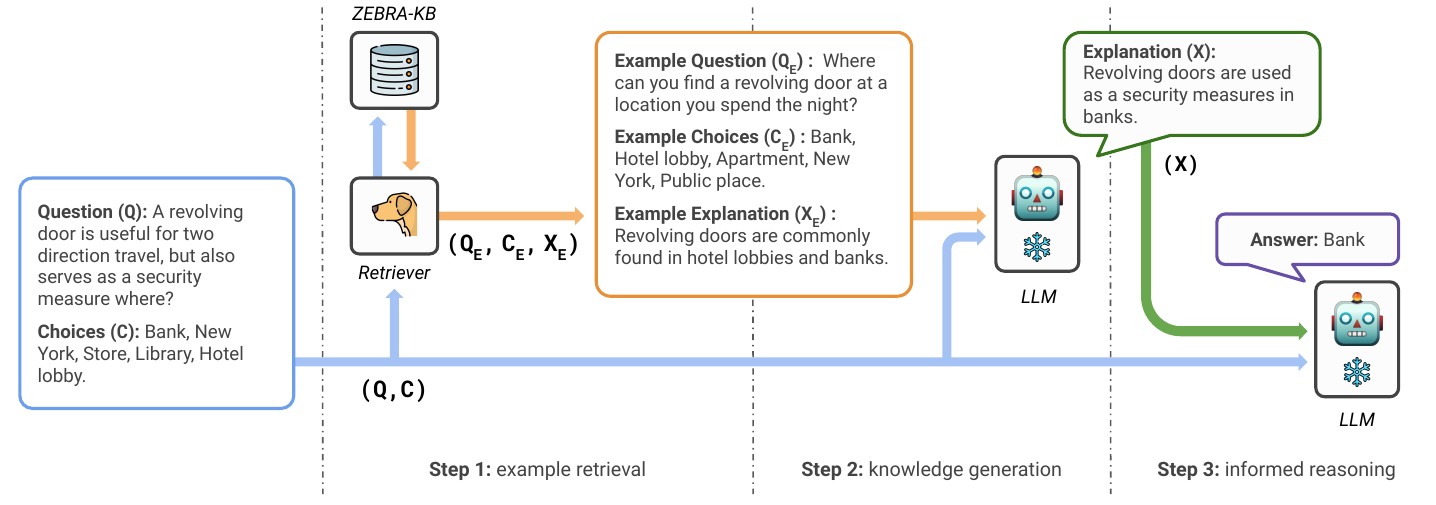}
    \caption{The \system{} framework in its entirety.
        Starting with a question $Q$ and its possible choices $C$, the first step (\textit{example retrieval}) is to ask the retriever to fetch relevant examples from a collection made of questions along with their choices and associated knowledge explanations $(Q_e, C_e, X_e)$.
        Then, the model is asked to generate one or more explanations $X$ for the question $Q$ with choices $C$ emulating the relationship in the elements $(Q_e, C_e, X_e)$ of the examples (\textit{knowledge generation} step).
        Finally, during the \textit{informed reasoning} step, the same model is asked to perform question answering on the question $Q$ given the choices $C$ and the generated knowledge explanations $X$.}
    \label{fig:zebra}
\end{figure*}

\paragraph*{Knowledge retrieval.}
There have also been several efforts to equip LLMs with mechanisms to retrieve commonsense knowledge from structured and unstructured data~\cite{rag-intensive}.
Noteworthy is the work by \citet{yu-etal-2022-retrieval}, who introduced RACo, a retriever-generator approach to retrieve and use commonsense knowledge in knowledge-intensive tasks.
More specifically, the knowledge is drawn by the retriever from a large knowledge base comprising twenty million commonsense statements, situations, and events collected from various resources and benchmarks, such as ConceptNet, ATOMIC, and GKB~\cite{bhakthavatsalam2020genericskb}, among others.
Then, the knowledge is integrated into a reader model via Fusion-in-Decoder~\cite[FiD]{izacard2020leveraging} that is trained to output the correct answer.
Unfortunately, these approaches suffer from the same drawbacks as the previous ones, as they require additional training and fine-tuning, and their performance is limited by the quality and completeness of the knowledge base used.

\paragraph*{Knowledge generation.}
\citet[Self-Talk]{shwartz-etal-2020-unsupervised} and \citet[GKP]{liu-etal-2022-generated} were the first to introduce approaches to generate commonsense knowledge about the input question before providing the answer.
However, the generalizability and application of such approaches to broader domains is hindered by the need for human intervention: Self-Talk requires hand-crafted templates specific to each dataset, whereas GKP requires human-written explanations designed specifically for each task.

Finally, introspection has been proposed to improve the interpretability of language models by generating contextually relevant knowledge for a given question.
\citet{liu-etal-2022-rainier} were the first to introduce the concept of \textit{knowledge introspection}, which paved the way to the development of introspective reasoners, including Rainier~\cite{liu-etal-2022-rainier} and Crystal~\cite{liu-etal-2023-crystal}.
These approaches are based on teaching an LLM to generate knowledge that is specifically helpful in answering a given question and, at the same time, in optimizing their generated answers conditioned on the knowledge generated during the first phase via reinforcement learning techniques.
Although introspective reasoners have achieved remarkable results, they still require additional training and fine-tuning; moreover, the generated knowledge and their reasoning patterns are strongly limited by what they have seen at training time.
In conclusion, to the best of our knowledge, no work has yet explored combining retrieval and introspection with a view to enhancing the reasoning capabilities of language models in a zero-shot setting without the need for additional training of the LLM.\@

\section{\system{}}\label{sec:zebra}
In this section, we introduce \system{}, our novel zero-shot example-based retrieval augmentation framework for commonsense reasoning that combines the benefits of knowledge retrieval and introspection while dropping the need for additional training of the LLM.\@
\system{}'s pipeline is composed of three main steps, as illustrated in Figure~\ref{fig:zebra}: i) example retrieval (Section~\ref{sec:retriever-zebra}), ii) example-guided knowledge generation (Section~\ref{sec:kg}), and iii) knowledge-informed reasoning (Section~\ref{sec:ir}).

\subsection{Example Retrieval}\label{sec:retriever-zebra}
The first step of \system{} is to retrieve the ``examples'' from a dedicated knowledge base.
The key difference from standard knowledge retrieval is that, instead of retrieving isolated facts or statements, we retrieve complete examples.
Each example consists of a question, a list of choices, and simple explanations that clarify how commonsense knowledge justifies the correctness or incorrectness of each choice in relation to the input question.\footnote{In the following, we use the terms ``explanations'' and ``knowledge'' interchangeably to refer to ``explanations that use commonsense knowledge.''}
Retrieving full examples allows \system{} to provide a broader context for the reasoning process that the LLM has to follow to generate the knowledge necessary to answer the question.

\paragraph*{Retriever architecture.}
Our example retriever builds on top of DPR \citep[Dense Passage Retrieval]{karpukhin-etal-2020-dense}, which uses an encoder to produce a dense representation of the query and the passages.
Given an input query $q$ and a passage $p \in \mathcal{P}$, where $\mathcal{P}$ is a collection of passages, the $\operatorname{Retriever(\cdot)}$ model computes the embeddings of $q$ and $p$:
\begin{equation*}
    E_Q(q) = \operatorname{Retriever}(q), \hspace{0.2em} E_P(p) = \operatorname{Retriever}(p)
\end{equation*}
Then, we rank the most relevant passages with respect to $q$ using the similarity function $\operatorname{sim}(q, p)=E_{Q}(q)^{\top} E_P(p)$, i.e., the dot product between the query and passage embeddings.
The retriever returns the top-$k$ passages $\mathcal{P}(q)$ that are most similar to the input query $q$.

\paragraph*{Query and passage representation.}
For our retriever to work, it is fundamental to encode the queries (input questions and their choices) and the passages (questions of the examples and the corresponding choices) in a way that allows the model to retrieve the most relevant ones, i.e., the positives, and tell them apart from the irrelevant ones, i.e., the negatives.
Let the query $q = (Q, C)$ be composed of a question $Q$ and a list of possible choices $C = (c^1, c^2, \dots, c^n)$.
Then, we represent the query $q$ as the concatenation of the question $Q$ and each choice $c^i$ separated by a special token:
\begin{equation}\label{eq:example}
        Q \: \text{[SEP]} \: c^1 \: \text{[SEP]} \: c^2 \: \dots \: \text{[SEP]} \: c^n
\end{equation}
For each query $q$, we sample a set of positive passages $\overline{\mathcal{P}}(q)$ from all the queries in the knowledge base of examples.
More specifically, a passage $p$ is considered the positive of a query $q$ if the two share the same main topic, i.e., they are about the same concept.
In addition, we augment the set of positive passages by permuting, removing, or reordering the choices in the queries, as the model should be able to retrieve relevant examples even when the choices are presented in a different order or present different distractors.

\paragraph*{Training objective.}
We train the retriever using multi-label noise contrastive estimation (NCE) as the training objective. The $\mathcal{L}_{Retriever}$ loss for $q$ is defined as:
\begin{equation} \label{eq:retriever_loss}
    -\log \sum_{p^{+} \in \overline{\mathcal{P}}(q)}\frac{e^{\operatorname{sim}\left(q, p^{+}\right)}}{e^{\operatorname{sim}\left(q, p^{+}\right)}+\sum_{p^{-}\in \widehat{\mathcal{P}}(q)} e^{\operatorname{sim}\left(q, p^{-}\right)}}
\end{equation}
where $\overline{\mathcal{P}}(q)$ are the positives for $q$, and $\widehat{\mathcal{P}}(q)$ is the set of negative samples for $q$, built using the positives of the other queries in the same batch.

\subsection{Example-Guided Knowledge Generation}\label{sec:kg}
Given a question $Q$ and a list of choices $C = (c^1, c^2, \dots, c^n)$, the next step in \system{} is to generate a list of explanations $X$ that can help in answering the question $Q$.
Unlike introspective reasoners, which generate relevant knowledge directly, we build on top of case-based reasoning.
Our approach encourages an LLM to generate knowledge by emulating the relationship in the question-knowledge pairs found in the retrieved examples. 
More specifically, we retrieve the top $k$ examples $E$ that are conceptually relevant to the given input $q$:
\begin{equation}
    E = \operatorname{top-k}(\operatorname{Retriever}(q))
\end{equation}
Each example $e_i \in E \; \forall i=1,\dots,k$ is composed of a question $Q_{e_i}$, choices $C_{e_{i}} = (c^1_{e_{i}}, c^2_{e_{i}}, \dots, c^n_{e_{i}})$ and a sequence of gold or silver explanations $X_{e_{i}} = (x^1_{e_{i}}, x^2_{e_{i}}, \dots, x^m_{e_{i}})$ that can help answering $Q_{e_i}$.
Subsequently, we construct a prompt containing all of these items of information and ask the LLM to generate a list of explanations $X$ for $(Q, C)$, following the relationship in the questions $Q_{{e}_{1:k}}$, choices $C_{{e}_{1:k}}$, and explanations $X_{{e}_{1:k}}$ of the top-$k$ retrieved examples:
\begin{equation}
    X = \operatorname{Prompt}(Q_{{e}_{1:k}}, C_{{e}_{1:k}}, X_{{e}_{1:k}}, Q, C)
\end{equation}
We provide more details about the construction of the prompt in Appendix~\ref{app:kg}.

\subsection{Knowledge-Informed Reasoning}\label{sec:ir}
Having generated the list of explanations $X$ for $(Q, C)$, \system{} proceeds to the final step, where the LLM is asked to perform question answering on the input question $Q$ with the list of choices $C$ and the explanations $X$.
Therefore, the model is asked to predict the correct answer $A$ by conditioning on the input $(Q, C, X)$:
\begin{equation}
    A = \operatorname{argmax}_{c \in C} P(c | Q, C, X)
\end{equation}
where $P(c | Q, C, X)$ is the probability of the choice $c$ given $Q$, $C$, and $X$.
In practice, we compute the probability of the label assigned by the model to each choice $c$ and select the one with the highest probability as the final prediction.
We note that this approach can easily be extended to a few-shot setting by providing the model with a few examples before asking the question $Q$.
We provide additional details about the prompt used for this step in Appendix \ref{app:qa}.

\subsection{\kb{}}\label{sec:zebra-kb}
\system{} requires a knowledge base of examples to retrieve and generate commonsense knowledge.
Specifically, each example in the knowledge base is composed of a question, a list of choices, and a list of explanations that can help answering the question.
These examples can be drawn from the training sets of well-established question answering datasets.
A dataset providing such examples is the CommonsenseQA dataset \cite{talmor-etal-2019-commonsenseqa}, which benefits from a manually-annotated explanations~\cite[CoS-E and ECQA]{rajani-etal-2019-explain,aggarwal-etal-2021-explanations}.
However, not every QA dataset provides such explanations.
To obtain an inexpensive but effective solution, we propose \kb{}, a new knowledge base of examples with gold and silver explanations, the latter being generated using commercially-available LLMs, such as GPT and Gemini.
We provide more details about the generation of \kb{} in Section \ref{sec:zebra-kg-generation} and Appendix~\ref{app:kb}.

\section{Experimental Setup}
In this section, we describe the experimental setup used to train and evaluate \system{}.
More specifically, we provide an overview on the training process of the retriever (Section~\ref{sec:retriever-exp}), the datasets used to evaluate our framework (Section~\ref{sec:datasets}), the process to generate \kb{} (Section~\ref{sec:zebra-kg-generation}), as well as the models that we consider for our experiments (Section~\ref{sec:models}).

\subsection{Retriever}\label{sec:retriever-exp}
We build our retriever on top of E5-base-v2 \cite{wang2022text}, a small-sized transformer-based encoder (about 109M parameters) that is pre-trained on a large corpus of text.
We fine-tune the retriever on the training set of CommonsenseQA~\cite[CSQA]{talmor-etal-2019-commonsenseqa} using the procedure described in Section~\ref{sec:retriever-zebra}.
In CSQA, each question is associated with a topic or concept, which we use to construct the set of positive examples for each query.
More specifically, we consider two questions $q$ and $q'$ in the dataset as pair-wise positives if they are tagged with the same topic or concept.
We select at most 64 positive examples for each query and use up to 200 negatives per batch.
We train the encoder for a maximum of 25,000 steps using RAdam~\cite{liu2019radam} with a learning rate of \texttt{1e-5} and a linear learning rate decay schedule.
At the end of the training, we select the best model based on the loss on the validation set of CSQA.
We highlight that the retriever is trained only once and is then used to retrieve examples for all the LLMs and datasets we evaluate.
Moreover, the retriever is the only trained component in \system{}, as the parameters of the LLMs are kept frozen during the entire process.

\subsection{Evaluation Benchmarks}\label{sec:datasets}
We evaluate our approach against 8 well-established QA datasets: CommonsenseQA~\cite[CSQA]{talmor-etal-2019-commonsenseqa}, OpenBookQA~\cite[OBQA]{mihaylov-etal-2018-suit}, ARC-Easy and ARC-Challenge~\cite{clark2018think}, PIQA~\cite{bisk2019piqa}, WinoGrande~\cite[WG]{sakaguchi2019winogrande}, CommonsenseQA 2.0~\cite[CSQA2]{talmor2022commonsenseqa} and QASC~\cite{khot2020qasc}.
To ensure fair and consistent comparisons with recent work~\cite{yu-etal-2022-retrieval,liu-etal-2023-crystal}, we follow standard practice by evaluating on test sets when their labels are publicly available, and otherwise on development sets. Specifically, we use the development sets for CSQA, PIQA, WG, CSQA2, and QASC, and the test sets for ARC and OBQA.

\begin{table}[t]
    \centering
    \adjustbox{max width=\columnwidth}{
    \begin{tabular}{l|l}
        \toprule
        \textbf{HuggingFace model ID} & \textbf{Alias} \\ \midrule
        \texttt{mistralai/Mistral-7B-Instruct-v0.2} & Mistral-v0.2 \\ 
        \texttt{microsoft/Phi-3-small-8k-instruct} & Phi-3-Small\\ 
        \texttt{meta-llama/Meta-Llama-3-8B-Instruct} & Llama-3 \\ 
        \texttt{microsoft/Phi-3-mini-128k-instruct} & Phi-3-Mini \\ \bottomrule
    \end{tabular}
    }
    \caption{List of LLMs considered for \system{}. For better readability, we provide a mapping from the original HuggingFace model IDs to their aliases.}
    \label{tab:models}
\end{table}

\begin{table*}[htbp!]
    \centering
    \begin{adjustbox}{max width=0.95\textwidth}
    \centering
        \begin{tabular*}{\textwidth}{@{\extracolsep{\fill}}clccccccc}
            \toprule
                                                      &              &      & \multicolumn{5}{c}{\textbf{ZEBRA}} &                                                                                          \\

            \cmidrule{4-8}

            \textbf{Size}                            & \textbf{Model}       &      & \textbf{k = 1}                     & \textbf{k = 3}   & \textbf{k = 5} & \textbf{k = 10}  & \textbf{k = 20} & \textbf{Oracle} \\
            \midrule
            \multirow{3}{*}{\textit{7 $\sim$ 8B}} & Mistral-v0.2         & 68.2 & 68.6                               & \underline{72.5} & \textbf{73.3}  & 72.1             & 71.8            & 90.3            \\
                                                     & Phi-3-Small          & 77.2 & \underline{80.8}                   & 80.7             & \textbf{80.9}  & 79.5             & 79.6            & 95.2            \\
                                                     & Llama-3             & 73.9 & 77.4                               & \textbf{78.7}    & \textbf{78.7}  & \underline{78.0} & 76.6            & 95.5            \\
            \midrule
            \multirow{1}{*}{\textit{$\sim$ 4B}} & Phi-3-Mini           & 73.4 & \textbf{75.3}                   & \underline{74.9} & 74.8           & 73.9             & 72.8            & 94.9            \\
            \midrule
                                                     & Average              & 73.2 & 75.5                               & \underline{76.7} & \textbf{76.9}  & 75.9             & 75.2            & 94.0            \\
                                                     & $\Delta$ Improvement & --   & +2.3                               & \underline{+3.5} & \textbf{+3.7}  & +2.7             & +2.0            & +20.8           \\
            \bottomrule
        \end{tabular*}
    \end{adjustbox}
    \caption{Results in terms of accuracy on the CSQA development set. Here, $k$ is the number of examples used during the knowledge generation step. ``Oracle'' indicates the results of the models when they have access to the manually-created explanations of ECQA. Best results are in \textbf{bold}, while second-best results are \underline{underlined}.}
    \label{tab:csqa}
\end{table*}

\subsection{Creating \kb{}}\label{sec:zebra-kg-generation}
CommonsenseQA features a manually annotated set of explanations for each question that was originally introduced in prior work \cite[ECQA]{aggarwal-etal-2021-explanations}.
However, this is not a realistic scenario for QA datasets in general and for real-world applications.
As a matter of fact, CSQA and OBQA are the only datasets in our evaluation that provide such explanations.
However, this limitation does not prevent us from evaluating our framework on the other datasets, as we can generate the required explanations using a silver annotation procedure that is effective and inexpensive.

To create our silver explanations we start from the training set of each dataset and, for each sample, we consider its question $Q$, choices $C$, and -- most importantly -- the correct answer $A$.
Then, we use Google's GenerativeAPI to prompt Gemini-1.5-Flash\footnote{All our explanations are generated using the latest version of Gemini-1.5-Flash available in May 2024.} to generate a list of explanations given $Q$, $C$, and $A$, setting the temperature to \textit{0.0} (for reproducibility) and the maximum number of new tokens to \textit{256}.
We select the top-10 explanations returned by Gemini.
The result is a knowledge base of exemplar commonsense knowledge, which we refer to as \kb{}.
We note that relying on commercially-available LLMs limits their output, i.e., the generated explanations, when the questions and/or the possible choices concern sensitive and possibly unsafe topics, e.g., drugs, sex, violence, and race, among others \cite{tedeschi2024alert}.
Future work may address this limitation in order to generate more comprehensive and diverse explanations. We provide the details about the prompt used for the generation of \kb{} in Appendix~\ref{app:kb}.

\begin{table*}[htbp!]
    \centering
    \resizebox{\textwidth}{!}{%
        \begin{tabular}{lccccccc|c|c}
            \toprule
            \textbf{Model} & \textbf{ARC-C}       & \textbf{ARC-E}       & \textbf{OBQA}        & \textbf{PIQA}        & \textbf{WG}  & \textbf{CSQA2}       & \textbf{QASC}        & \textbf{Avg.}  & \textbf{$\Delta$}     \\
            \midrule
            Mistral-v0.2  & 72.4 / \textbf{75.2} & 85.8 / \textbf{87.4} & 68.8 / \textbf{75.8} & 76.1 / \textbf{80.2} & 55.8 / \textbf{60.7} & 58.5 / \textbf{67.5} & 66.1 / \textbf{68.3} & 69.1 / \textbf{73.6}  & \textbf{+4.5}  \\
            Phi-3-Small   & 90.4 / \textbf{91.6} & 96.9 / \textbf{97.7} & 90.4 / \textbf{91.2} & 86.6 / \textbf{88.1} & 79.1 / \textbf{81.0} & 68.0 / \textbf{74.6} & \textbf{83.5} / 81.0 & 85.0 / \textbf{86.4}  & \textbf{+1.4}  \\
            Llama-3       & 79.4 / \textbf{83.5} & 91.7 / \textbf{92.9} & 73.4 / \textbf{79.6} & 78.3 / \textbf{84.0} & 56.2 / \textbf{63.2} & 64.3 / \textbf{69.4} & 78.2 / \textbf{79.1} & 74.5 / \textbf{78.8}  & \textbf{+4.3}  \\
            Phi-3-Mini    & 85.7 / \textbf{88.0} & 95.4 / \textbf{96.0} & 82.8 / \textbf{87.8} & 80.4 / \textbf{84.2} & 67.3 / \textbf{72.9} & 59.3 / \textbf{64.6} & \textbf{74.7} / 73.9 & 77.9 / \textbf{81.0}  & \textbf{+3.1}  \\
            \bottomrule
        \end{tabular}%
    }
    \caption{Results in terms of accuracy on 7 commonsense benchmarks: ARC-Challenge (ARC-C), ARC-Easy (ARC-E), OpenBookQA (OBQA), PIQA, WinoGrande (WG), CommonsenseQA 2.0 (CSQA2), and QASC. The results are reported in the format \textit{zero-shot} / \system{} with $k = 5$ retrieved examples. Best results are in \textbf{bold}.}
    \label{tab:all-datasets}
\end{table*}

\subsection{Models}\label{sec:models}
We evaluate \system{} using four instruction-tuned LLMs. Table \ref{tab:models} presents the models selected for evaluation and provides a mapping from their original HuggingFace model IDs to the aliases used in this paper for better readability.
Three of these models -- Mistral-v0.2, Phi-3-Small, and Llama-3 -- feature a similar number of parameters.
We also include Phi-3-Mini to evaluate the effectiveness of \system{} on a LLM with a significantly lower number of parameters (3.8 billion compared to 7--8 billion of the others).
This set of LLMs allows us to evaluate our approach on top of the strongest LLMs available at the time of writing, as well as on smaller models that are computationally less expensive.

\section{Results}\label{sec:results}
In this section, we present and discuss the results of \system{} on the 8 commonsense question-answering benchmarks introduced in Section~\ref{sec:datasets} using the models listed in Section~\ref{sec:models}.

\subsection{Results on CSQA}\label{sec:quantitative-csqa}
Table~\ref{tab:csqa} provides an overview of the results obtained by different LLMs on CSQA~\cite{talmor-etal-2019-commonsenseqa}.
For each model, we report the score obtained in the zero-shot setting, as well as the performance when using the \system{} framework with different numbers of retrieved examples $k$.
Note that, here, $k$ is the number of examples used during the knowledge generation step, not the number of in-context examples used for question answering.
We also report the ``oracle'' accuracy of each model when using the manually-created explanations provided in ECQA (right-most column in Table~\ref{tab:csqa}), which should represent the upper bound of the performance for an LLM when a human provides one or more explanations to a question having the correct answer available.
It is important to underline that, because our retriever model is trained specifically on the CSQA training set, these results can also be viewed as an extrinsic evaluation of the in-domain performance of the retriever.

We can immediately see that \system{} consistently improves the performance of all the LLMs, with an average increase of 3.7 points of accuracy when $k=5$.
Moreover, we can observe that the performance gain is approximately the same across different LLMs, independently of the architecture and the number of parameters, highlighting the reliability of our approach.
Although the results obtained with the oracle knowledge seem to suggest an even higher increase in performance, we stress the fact that those explanations often contain an explicit link between the question and the correct answer, making the models more likely to take a ``shortcut'' to the correct answer.
Indeed, as mentioned above, the explanations are hand-crafted by a human who has access not only to the question and the choices but also to the correct answer, which is not the case in a real-world scenario.
We provide examples of this pattern in Appendix~\ref{app:ecqa}.

\begin{figure*}[t]
    \centering
    \begin{subfigure}[t]{0.24\textwidth}
        \centering
        \includegraphics[width=\textwidth]{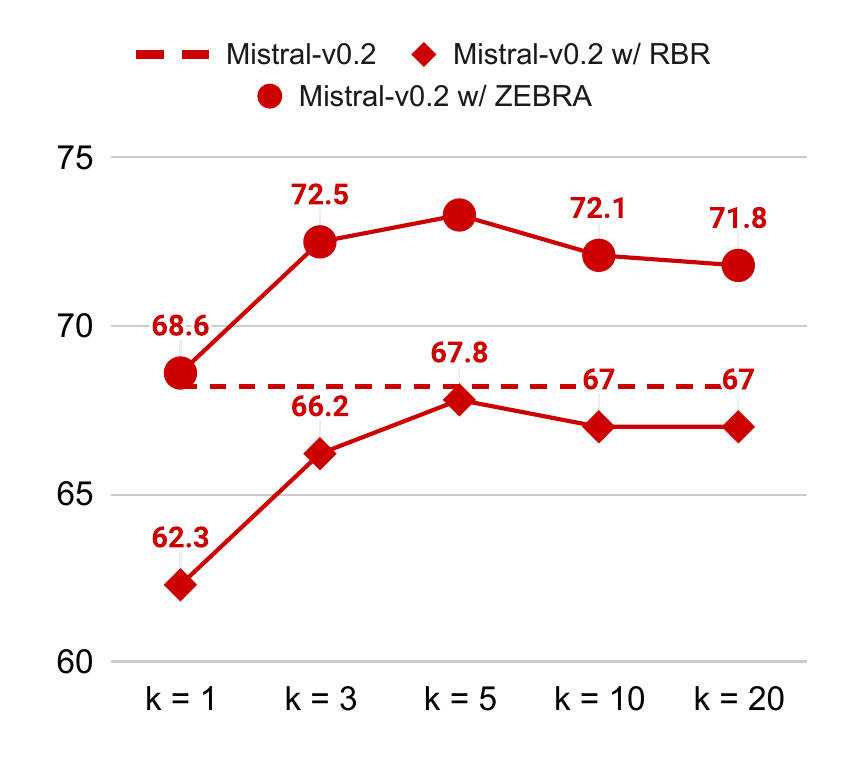}
        \caption{Mistral-v0.2}
        \label{fig:chart1}
    \end{subfigure}
    \begin{subfigure}[t]{0.24\textwidth}
        \centering
        \includegraphics[width=\textwidth]{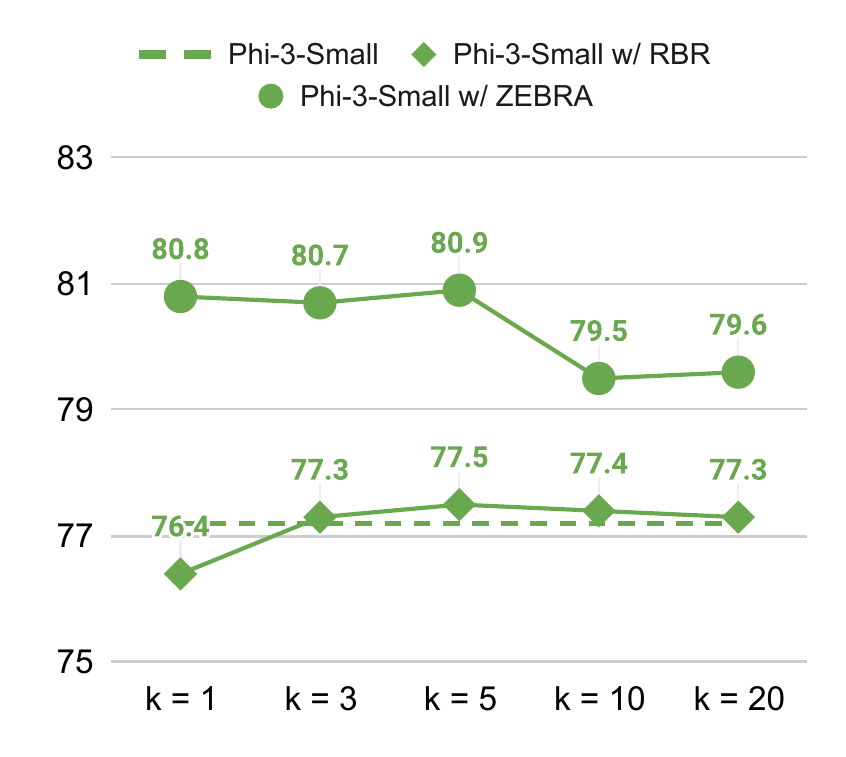}
        \caption{Phi-3-Small}
        \label{fig:chart2}
    \end{subfigure}
    \begin{subfigure}[t]{0.24\textwidth}
        \centering
        \includegraphics[width=\textwidth]{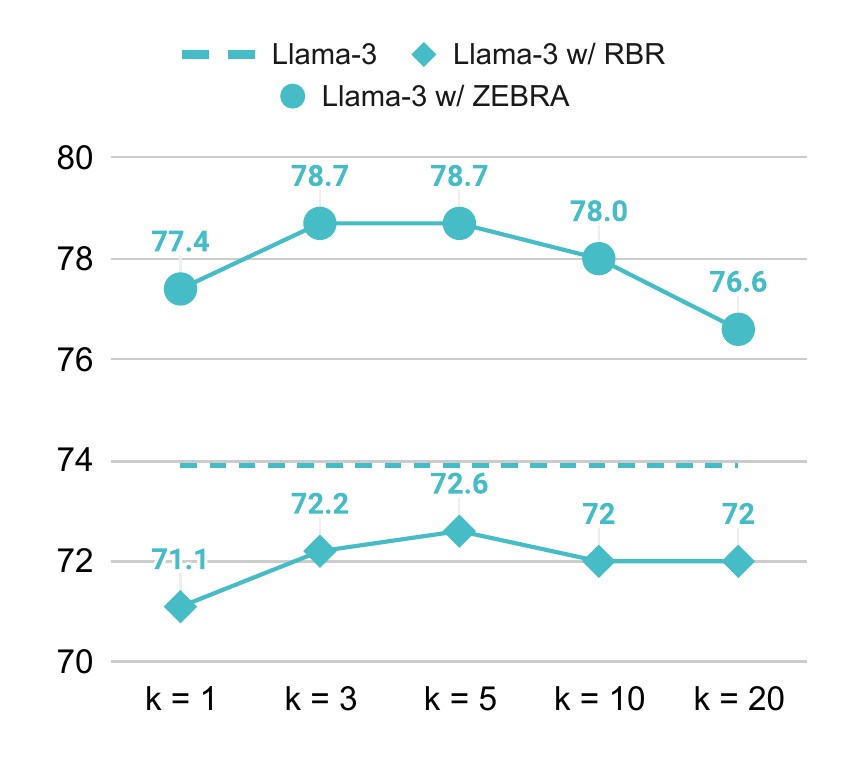}
        \caption{Llama-3}
        \label{fig:chart3}
    \end{subfigure}
    \begin{subfigure}[t]{0.24\textwidth}
        \centering
        \includegraphics[width=\textwidth]{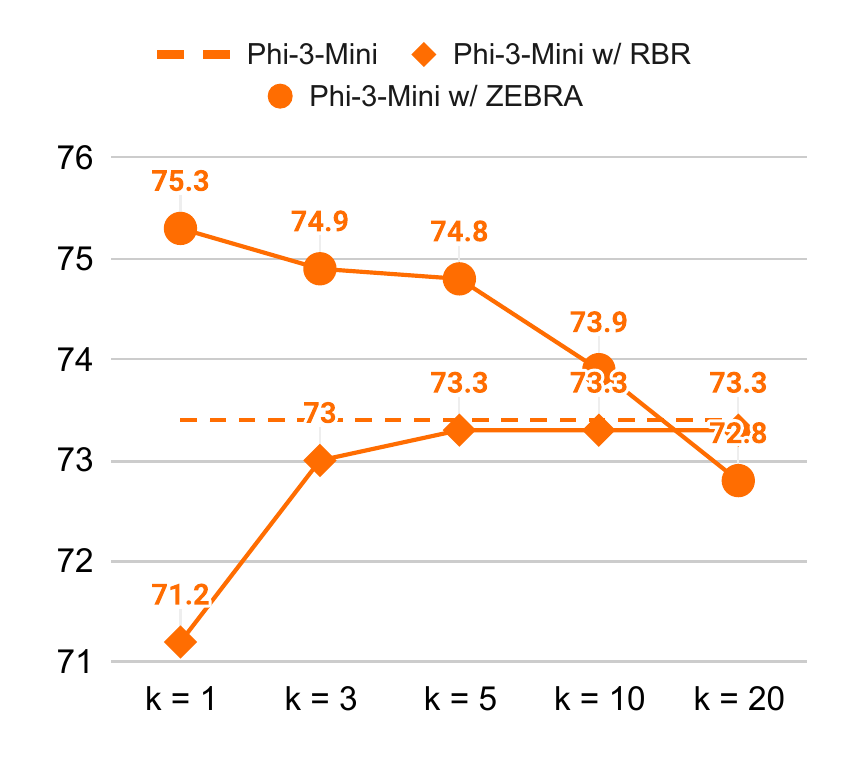}
        \caption{Phi-3-Mini}
        \label{fig:chart4}
    \end{subfigure}
    \caption{Comparison of the LLMs performance on the CSQA development set using \system{} and direct knowledge retrieval (RACo-based Retrieval) as the number of retrieved examples/knowledge statements $k$ increases.}
    \label{fig:four_charts}
\end{figure*}

\subsection{Results on Other Benchmarks}\label{sec:quantitative-all}
The results of \system{} on CSQA in Table~\ref{tab:csqa} help us determine the best number of examples to use during the knowledge generation step, which we find to be $k=5$.
Having established the best value for $k$, we shift our focus to the evaluation of the generalizability of \system{} on another 7 benchmarks, which assess different aspects of commonsense reasoning, e.g., physical interactions in PIQA, coreference resolution in WG, science knowledge in ARC and OBQA, etc.
Importantly, for each of these benchmarks, we retrieve the $k$ examples for each question from the corresponding training set contained in \kb{}.

Table~\ref{tab:all-datasets} summarizes the results, where we can see that \system{} consistently outperforms the baselines across all the 7 benchmarks and all the 4 LLMs that we consider, with the exception of QASC for Phi-3-Small and Phi-3-Mini.
The consistent improvement in performance across different datasets and LLMs highlights the reliability of \system{} even when the knowledge base is generated through a silver annotation procedure and even when the retriever is trained on a dataset (CSQA) which is potentially very different from the ones used in this evaluation.
In Appendix~\ref{app:ood}, we present additional experiments demonstrating that \system{} remains effective even when the retrieved examples do not come from the same distribution of the evaluation dataset. This highlights \system{}'s robustness when both the retriever and the knowledge base are affected by domain shift.

\begin{table*}[t]
    \centering
    \begin{adjustbox}{max width=\textwidth}
        \begin{tabular}{clcccccccccc|c}
            \toprule
                                                                      & \textbf{System} & \textbf{Generator} & \textbf{Reasoner} & \textbf{CSQA} & \textbf{ARC-C} & \textbf{ARC-E}
                                                                      & \textbf{OBQA} & \textbf{QASC} & \textbf{PIQA} & \textbf{CSQA2} & \textbf{WG} & \textbf{AVG}
            \\
            \midrule
            & Llama-3 & -- & Llama-3 & 73.9 & \underline{79.4} & \textbf{91.7}  & \underline{73.4} & 78.2 & \underline{78.3} & \underline{64.3} & 56.2 & \underline{74.4} \\
            \midrule
            \multirow{3}{*}{\rotatebox[origin=c]{90}{\scalebox{0.9}{\textit{Superv.}}}} & Rainier-large & T5 0.77B & Llama-3
                                                                      & \cellcolor{red!15}72.9
                                                                      & \cellcolor{red!15}76.0
                                                                      & \cellcolor{red!15}88.6
                                                                      & \cellcolor{red!15}71.4
                                                                      & \cellcolor{red!15}74.5
                                                                      & \cellcolor{red!15}76.6
                                                                      & \cellcolor{red!15}57.1
                                                                      & \cellcolor{green!15}59.3
                                                                      & \cellcolor{red!15}72.0
            \\
                                                                      & Crystal-3B & T5 3B & Llama-3
                                                                      & \cellcolor{red!15}72.6
                                                                      & \cellcolor{red!15}75.5
                                                                      & \cellcolor{red!15}89.5
                                                                      & \cellcolor{red!15}72.6
                                                                      & \cellcolor{red!15}75.9
                                                                      & \cellcolor{red!15}77.7
                                                                      & \cellcolor{red!15}58.6
                                                                      & \cellcolor{green!15}60.1
                                                                      & \cellcolor{red!15}72.8
            \\
                                                                      & Crystal-11B & T5 11B & Llama-3
                                                                      & \cellcolor{green!15}\underline{75.1}
                                                                      & \cellcolor{red!15}77.3
                                                                      & \cellcolor{red!15}91.2
                                                                      & \cellcolor{red!15}72.6
                                                                      & \cellcolor{green!15}\underline{78.4}
                                                                      & \cellcolor{red!15}78.2
                                                                      & \cellcolor{red!15}60.0
                                                                      & \cellcolor{green!15}\underline{60.5}
                                                                      & \cellcolor{red!15}74.1
            \\
            \midrule
            \multirow{3}{*}{\rotatebox[origin=c]{90}{\scalebox{0.9}{\textit{Unsuperv.}}}} & Self-Talk & Llama-3 & Llama-3
                                                                      & \cellcolor{red!15}70.6
                                                                      & \cellcolor{red!15}~~78.7*
                                                                      & \cellcolor{red!15}~~91.4*
                                                                      & \cellcolor{red!15}~~72.2*
                                                                      & \cellcolor{green!15}~~78.3*
                                                                      & \cellcolor{red!15}77.2
                                                                      & \cellcolor{red!15}~~63.4*
                                                                      & \cellcolor{green!15}58.3
                                                                      & \cellcolor{red!15}73.8
            \\
                                                                      & GKP & Llama-3 & Llama-3
                                                                      & \cellcolor{green!15}74.0
                                                                      & \cellcolor{red!15}~~78.5*
                                                                      & \cellcolor{red!15}~~\underline{91.5}*
                                                                      & \cellcolor{red!15}~~70.0*
                                                                      & \cellcolor{red!15}76.9
                                                                      & \cellcolor{red!15}~~76.5*
                                                                      & \cellcolor{green!15}\textbf{65.9}
                                                                      & \cellcolor{green!15}~~60.4*
                                                                      & \cellcolor{red!15}74.2
            \\
                                                                      & ZEBRA \textbf{(Ours)} & Llama-3 & Llama-3
                                                                      & \cellcolor{green!15}\textbf{78.7}
                                                                      & \cellcolor{green!15}\textbf{84.3}
                                                                      & \cellcolor{red!15}90.9
                                                                      & \cellcolor{green!15}\textbf{80.0}
                                                                      & \cellcolor{green!15}\textbf{79.1}
                                                                      & \cellcolor{green!15}\textbf{84.0}
                                                                      & \cellcolor{red!15}63.2
                                                                      & \cellcolor{green!15}\textbf{69.4}
                                                                      & \cellcolor{green!15}\textbf{78.7}
            \\
            \bottomrule
        \end{tabular}
    \end{adjustbox}
    \caption{Accuracy scores on the CSQA, ARC-Challenge, ARC-Easy, OBQA, QASC, PIQA, CSQA2 and WG benchmarks when using \system{} compared to the baselines. Best results are in \textbf{bold}, while second-best results are \underline{underlined}. A green cell indicates an improvement in performance compared to Llama-3 without generated knowledge, while a red cell indicates a decrease. *: results computed using the original methodology on datasets not evaluated by the baseline authors.}
    \label{tab:crystal}
\end{table*}

\subsection{Comparison with Knowledge Retrieval}\label{sec:zebra-raco}
To further show the effectiveness of \system{}, we carry out a 1-to-1 comparison against an approach that retrieves commonsense knowledge statements directly.
Specifically, we train a retriever to fetch commonsense knowledge statements (rather than full examples) that are relevant to the input question using the same retrieval strategy as that introduced in RACo \cite{yu-etal-2022-retrieval}.
Moreover, to ensure a fair comparison, we replace the specialized reader in RACo -- a T5 model trained using FiD \cite{izacard2020leveraging} -- with the same LLMs we use in our experimental setup.
We refer to this approach as RACo-based Retrieval (RBR).

Figure~\ref{fig:four_charts} compares the results of the four LLMs that we consider in this work when using \system{} and RBR on the CSQA development set.
Here, we can see that \system{} consistently outperforms RBR across all the LLMs and all the values of $k$ with the exception of Phi-3-Mini when $k=20$.
Interestingly, standard knowledge retrieval often leads to a negative impact on the performance of the LLMs, as we can see in the case of Mistral-v0.2, Llama-3 and Phi-3-Mini, where we observe a decrease in performance when using RBR compared to the vanilla LLMs.
Our analysis highlights the limitations of current commonsense knowledge bases and underscores the need for explanations derived from a reasoning process specific to the input question, rather than relying on general commonsense facts, which may not be sufficient to solve the task effectively.

\subsection{Comparison with Knowledge Generation}\label{sec:zebra-crystal}
One important aspect of our approach is the manner in which the knowledge is generated. Therefore, here we compare the quality of the knowledge generated by \system{} against previous methods: two unsupervised -- namely, Self-Talk \cite{shwartz-etal-2020-unsupervised} and GKP \cite{liu-etal-2022-generated} -- and two supervised (introspective reasoners) -- namely, Rainier \cite{liu-etal-2022-rainier} and Crystal~\cite{liu-etal-2023-crystal}.

To test the quality of the generated knowledge of each system in a fair setting, we evaluate the different knowledge \textit{generators} (i.e., the module of the system that generates the knowledge) using the same LLM as a \textit{reasoner} (i.e., the module of the system that answers the question) for all the systems.
This allows us to evaluate whether \system{}'s example-based retrieval augmentation framework produces knowledge that is of higher quality compared to the hand-crafted templates in Self-Talk and the manually-curated explanations in GKP.
Moreover, this setting enables a direct comparison between \system{}, which does not require fine-tuning of the underlying LLM, and models that are specifically trained to generate relevant knowledge, such as Rainier and Crystal.

Table~\ref{tab:crystal} shows the accuracy scores of \system{} and all the baselines over the 8 benchmarks introduced in Section \ref{sec:datasets}.
Interestingly, we observe that most systems -- except for \system{} -- encounter difficulties in scoring higher than a vanilla Llama-3 model in a consistent way across the datasets  without any input knowledge (first row in Table~\ref{tab:crystal}).
These results suggest that current approaches are not suitable for improving the reasoning capabilities of current LLMs.
Instead, \system{} surpasses the best performing baseline by an average of 4.3 points of accuracy, showcasing the effectiveness of retrieving relevant examples for the input question and allowing the model to generate knowledge by mimicking the relationship in the retrieved question-knowledge pairs.
Notably, \system{} is also able to outperform supervised techniques that have been explicitly trained to generate and leverage knowledge when answering an input question.\footnote{For example, the original reasoner \cite[UnifiedQA]{khashabi-etal-2020-unifiedqa} in Rainier is trained for question answering but kept frozen during the training of the knowledge generator, whereas Crystal is jointly trained on question answering and knowledge generation using reinforcement learning.}
These results highlight the effectiveness of \system{} in providing a simple and effective framework to improve the performance of current LLMs on commonsense reasoning tasks.

\subsection{Human Evaluation}\label{sec:qualitative}
Besides better results on standard benchmarks, one of the most important strengths of \system{} is the interpretability of the answers provided by the LLMs.
Indeed, the knowledge generated by the LLMs can be used by humans to understand the reasoning process that led to the final answer.
To evaluate this aspect, we conduct a small-scale manual analysis on the quality of the knowledge generated by the LLMs.
We randomly sample 100 instances from the CSQA development set and ask three annotators to validate the quality of the knowledge generated by \system{} when using Llama-3.
The annotation process is conducted in a blind fashion, i.e., each annotator is not aware of the labels assigned by the other annotators.

Following standard practice in the field \cite{liu-etal-2022-generated}, each human annotator is assigned the task of evaluating the quality of the generated knowledge in relation to the input question, its answer choices, and the correct answer.
The assessment is carried out using three metrics:
\begin{itemize}
    \item \textbf{Relevance}: whether the generated knowledge is relevant to the topic or concepts mentioned in the question; relevant knowledge is not necessarily factual or helpful.
    \item \textbf{Factuality}: whether the generated knowledge is factually correct; factual knowledge is not necessarily relevant or helpful.
    \item \textbf{Helpfulness}: whether the generated knowledge helps in answering the question in a direct or indirect way. We distinguish between three categories: helpful (i.e., supports the correct answer), harmful (i.e., negates the correct answer or supports an incorrect answer), or neutral (neither helpful nor harmful).
\end{itemize}
Note that an item of helpful knowledge may be factually incorrect, hence all the three dimensions are important for a comprehensive evaluation.

Our human evaluation shows a strong consensus among annotators regarding the quality of the knowledge generated by \system{} using Llama-3. Specifically, all three annotators labeled 96 out of 100 instances as ``relevant'', 88 out of 100 instances as ``factual'', and 74 out of 100 instances as ``helpful''.
Moreover, in only 13 out of 100 cases was the knowledge deemed ``harmful'' by at least one annotator, while only 6 instances were tagged as ``harmful'' by all three annotators.
Finally, there was no instance which was tagged as ``not relevant, not factual and harmful'' by an annotator, showing the reliability of the explanations generated with \system{}.
For reference, in Appendix \ref{app:human} we provide a number of qualitative examples of the knowledge generated by \system{} when using Llama-3 over the CSQA development set.

\section{Conclusions}
In this paper we presented \system{}, a novel zero-shot example-based retrieval augmentation framework for commonsense reasoning.
\system{} is composed of two main components: an example retriever that fetches examples that are relevant to the input question from a dedicated knowledge base, and an LLM that generates relevant knowledge for the input question by following the relationship in the questions and knowledge provided in the examples.
This allows \system{} to tackle the limitations of previous methods such as commonsense knowledge retrievers, unsupervised methods relying on human intervention and introspective reasoners, providing a simple and effective way to improve the performance of LLMs on commonsense reasoning and question answering tasks.
Since a knowledge base of curated examples may not always be available, we also introduced \kb{}, a novel knowledge base of examples with silver knowledge explanations for each question, showing how \system{} can be used in conjunction with \kb{} to achieve strong zero-shot results on 8 well-established QA datasets for commonsense reasoning.
Our experiments support the effectiveness of \system{} over other state-of-the-art methods for knowledge augmented reasoning, such as specialized commonsense retrievers and introspective reasoners.
Finally, we investigated the interpretability of the answers provided by the LLMs through a human evaluation of the knowledge generated by \system{}. The results demonstrate that the knowledge is often perceived as reasonable by humans, aiding their understanding of the model's reasoning process leading to the final answer.

\section*{Limitations}
\system{} is not without its limitations. There are several aspects that could be improved in future work.
Here, we list and briefly discuss some of the main limitations of our current approach, which may be addressed in future work.

\paragraph{Retriever Performance:} The performance of the retriever is crucial for the overall performance of \system{}. In this work, our training strategy for the retriever is based on a contrastive learning approach that mainly identifies the positive for a question based on whether the two questions $q$ and $q'$ share or ``talk'' about the same concept.
Despite this approach showing itself to be effective in our experiments, it may not be the most effective strategy for all types of questions.
Our positive identification and negative mining strategies are similar to what is commonly used in the literature for dense retrieval, nevertheless, future work may focus on developing more effective and efficient retrieval methods for commonsense reasoning.

\paragraph{Knowledge Base:} The quality of the knowledge base is crucial for the overall performance of \system{}.
In this work, we overcome the lack of manually-crafted explanations for the training sets of many of the evaluation benchmarks we use in our evaluation by using \kb{}, a knowledge base of examples with silver knowledge explanations created through Google's Gemini-1.5-Flash, which is relatively fast and inexpensive to interrogate.
Future work may focus on the generation of better explanations by using more advanced cleaning and filtering techniques.

\paragraph{Multilinguality and cross-linguality:} Our approach is currently limited to English language benchmarks.
Future work may focus on extending the framework to multilingual and cross-lingual settings, where the knowledge base is composed of examples in multiple languages, and the LLMs are able to generate knowledge in a language different from the input question, as well as being able to handle input questions written in other languages.
This would be interesting to investigate, as the language diversity of the examples in the knowledge base may help or penalize the LLMs on different languages and improve the overall performance of the framework.

\section*{Acknowledgements}
\begin{center}
\noindent
    \begin{minipage}{0.1\linewidth}
        \begin{center}
            \includegraphics[scale=0.05]{fair.pdf}
        \end{center}
    \end{minipage}
    \hspace{0.01\linewidth}
    \begin{minipage}{0.70\linewidth}
        Simone Conia gratefully acknowledges the support of the PNRR MUR project PE0000013-FAIR, which fully funds his fellowship.
    \end{minipage}
    \hspace{0.01\linewidth}
    \begin{minipage}{0.1\linewidth}
        \begin{center}
            \includegraphics[scale=0.08]{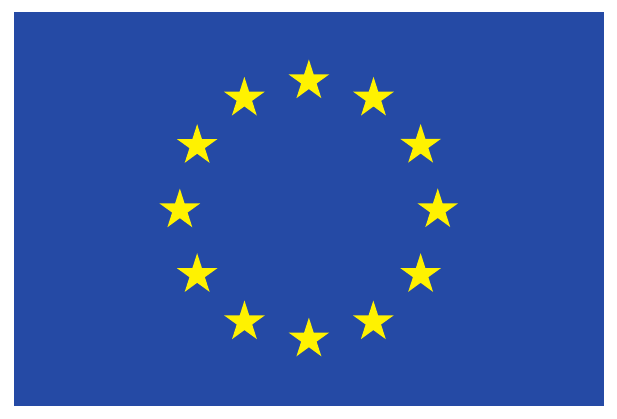}
        \end{center}
    \end{minipage}\\
\end{center}
\vspace{0.2cm}

\noindent Roberto Navigli also gratefully acknowledges the support of the CREATIVE project (CRoss-modal understanding and gEnerATIon of Visual and tExtual content), which is funded by the MUR Progetti di Rilevante Interesse Nazionale programme (PRIN 2020).

\bibliography{ms}

\appendix
\section{Appendix}

\subsection{Knowledge Generation Prompts}\label{app:kg}

In Table \ref{tab:prompt-kg} we report the prompt used for the knowledge generation step. The number of choices specified under the \textit{System} tag is dynamically adapted according to the number of choices available in the QA dataset.

\begin{table}[h]
    \renewcommand{\arraystretch}{1.6}
    \centering
    \begin{tabular}{|p{0.95\linewidth}|}
        \hline
        \textbf{System:} \textit{You are given a question and 5 choices}.

        \textit{Your task is to write one or more explanations that support the most likely option.}

        \textit{Note that:}

        * \textit{there is always one option that is correct and more likely than the others.}

        * \textit{the explanations must support only the most likely option and refute all the others.}

        * \textit{the explanations must be simple and concise (max 15 words).}

        \textit{Do you understand the task?}                                                                  \\
        \hline
        \textbf{Assistant:} \textit{Yes, I understand. Please provide the question and the possible choices.} \\
        \hline
        \textbf{User:}

        \textit{Question:}

        \{question\}
        \\
        \textit{Choices}:

        \{choices\}
        \\
        \hline
        \textbf{Assistant:} \textit{List of knowledge}:                                                       \\
        \hline
    \end{tabular}
    \caption{Prompt for the knowledge generation step.}
    \label{tab:prompt-kg}
\end{table}

\subsection{Question Answering Prompts}\label{app:qa}
In Table \ref{tab:prompt-mcq} we report the prompt used for the question answering task without additional input knowledge, which is the one used for the evaluation of the vanilla LLMs. The number and the list of labels appearing under the \textit{System} tag are adjusted according to the number of choices available in each QA dataset. Moreover, in Table \ref{tab:prompt-ir} we also report the prompt for the informed reasoning step, which is the exact same prompt as the one used for question answering, but with additional knowledge included. This latter can come from the knowledge generation step, from retrieval methods such as RACo, or from manual annotations such as ECQA.

\begin{table}[t!]
    \renewcommand{\arraystretch}{1.6}
    \centering
    \begin{tabular}{|p{0.95\linewidth}|}
        \hline
        \textbf{System:} \textit{You are a helpful assistant for question answering.}

        \textit{You are given a question and 5 choices (labeled A, B, C, D and E).}

        \textit{Your task is to choose the label corresponding to the best answer for the question.}

        \textit{Do you understand the task?}                                                                  \\
        \hline
        \textbf{Assistant:} \textit{Yes, I understand. Please provide the question and the possible choices.} \\
        \hline
        \textbf{User:}

        \textit{Question:}

        \{question\}
        \\
        \textit{Choices}:

        \{choices\}
        \\
        \hline
        \textbf{Assistant:} \textit{Answer}:                                                                  \\
        \hline
    \end{tabular}
    \caption{Prompt for the question answering task.}
    \vspace{-1.5mm}
    \label{tab:prompt-mcq}
\end{table}

\begin{table}[t!]
    \renewcommand{\arraystretch}{1.6}
    \centering
    \begin{tabular}{|p{0.95\linewidth}|}
        \hline
        \textbf{System:} \textit{You are a helpful assistant for question answering.}

        \textit{You are given a question, 5 choices (labeled A, B, C, D and E) and a list of explanations.}

        \textit{Your task is to choose the label corresponding to the best answer for the question based on the given explanations.}

        \textit{Do you understand the task?}                                                                  \\
        \hline
        \textbf{Assistant:} \textit{Yes, I understand. Please provide the question and the possible choices.} \\
        \hline
        \textbf{User:}

        \textit{Question:}

        \{question\}
        \\
        \textit{Choices}:

        \{choices\}
        \\

        \textit{Explanations}

        \{knowledge\}
        \\
        \hline
        \textbf{Assistant:} \textit{Answer}:                                                                  \\
        \hline
    \end{tabular}
    \caption{Prompt for the informed reasoning step.}
    \label{tab:prompt-ir}
\end{table}

\subsection{\kb{} Generation}\label{app:kb}
In Table \ref{tab:prompt-gemini} we show the prompt used to generate the silver knowledge using the Google GenerativeAPI and the Gemini-1.5-Flash large language model. The number and the list of labels appearing under the \textit{System} tag are adjusted according to the number of choices available in each QA dataset.

\begin{table}[t!]
    \renewcommand{\arraystretch}{1.6}
    \centering
    \begin{tabular}{|p{0.95\linewidth}|}
        \hline
        \textbf{System:} \textit{You are a helpful assistant for question answering.}

        \textit{You are given a question requiring commonsense knowledge to be solved, together with three possible choices (labeled A, B and C) and the label corresponding to the correct answer.}

        \textit{For each choice, generate a sentence with explicit commonsense knowledge that supports or refutes the choice.}

        \textit{The format of the generated knowledge should be in the following form:}

        \textit{A. ...}

        \textit{B. ...}

        \textit{C. ...} \\
        \hline
        \textbf{User:}

        \textit{Question:}

        \{question\}
        \\
        \textit{Choices}:

        \{choices\}
        \\
        \hline
    \end{tabular}
    \caption{Prompt for the generation of the silver knowledge given a question and its choices.}
    \label{tab:prompt-gemini}
\end{table}

\subsection{ECQA Explanations}\label{app:ecqa}
In Table \ref{tab:example-shortcut} we report five examples of instances coming from the development set of the CommonsenseQA dataset \cite[CSQA]{talmor-etal-2019-commonsenseqa}, which were manually annotated with a list of explanations by \citet{aggarwal-etal-2021-explanations} (ECQA). From the table we can see how the explanations contain an explicit link between the question and the correct answer, like \textit{What would go on top of wood?} where the correct answer is \textit{carpet} and the explanation is \textit{It is the carpet that could go on top of wood}. If these explanations were to be used as oracle knowledge during the informed reasoning step, the model would likely exploit the information to select the correct answer, resulting in a high probability of success. Consequently, although the results in Table \ref{tab:csqa} under the oracle column exhibit a significant performance increase compared to both the baselines and \system{}, we contend that this improvement is predominantly attributable to the models leveraging this shortcut.

\begin{table*}[t!]
    \begin{tabular}{>{\centering\arraybackslash}m{0.35\linewidth}>{\centering\arraybackslash}m{0.35\linewidth}>{\centering\arraybackslash}m{0.2\linewidth}}
        \toprule
        \textbf{Question}                                                                          & \textbf{Oracle Knowledge}                                                                                               & \textbf{Choices}                                                                                                                              \\
        \midrule
        The man often made smart remarks, like that any restaurant is a Mexican restaurant where?  & Mexican restaurants are found in Mexico. Mexico has many Mexican places.                                                & \begin{enumerate}[A.] \itemsep-0.3em \item city \item mexica \item san diego \item spain \item \textbf{mexico} \end{enumerate}                \\
        \midrule
        The man in the white suit was very lazy.  He did nothing useful.  Meanwhile, the man in the blue suit had put in effort and was very what?
        & The man in the white suit was very lazy. He did nothing useful. Meanwhile, the man in the blue had put in effort and was very productive. 
        & \begin{enumerate}[A.] \itemsep-0.3em \item restless \item active \item lazybutt \item \textbf{productive} \item hard work \end{enumerate} \\
        \midrule
        What could go on top of wood?                                                              & It is the carpet that could go on top of wood.                                                                          & \begin{enumerate}[A.] \itemsep-0.3em \item lumberyard \item synagogue \item floor \item \textbf{carpet} \item hardware store \end{enumerate}  \\
        \midrule
        Where could you find a toilet that only friends can use?                                   & Your friends come to your apartment. A toilet your apartment can only be used by your friends.                          & \begin{enumerate}[A.] \itemsep-0.3em \item rest area \item school \item stadium \item \textbf{apartment} \item hospital \end{enumerate}       \\
        \midrule
        The weasel was becoming a problem, it kept getting into the chicken eggs kept in the what? & The weasel was becoming a problem at the barn. The chicken eggs were kept into the barn and weasel was getting into it. & \begin{enumerate}[A.] \itemsep-0.3em \item forest \item \textbf{barn} \item public office \item out of doors \item freezer \end{enumerate}    \\
        \bottomrule
    \end{tabular}
    \caption{Examples of gold explanations from ECQA. The gold answers for the questions are in \textbf{bold}.}
    \label{tab:example-shortcut}
\end{table*}

\subsection{Out-of-domain Results}\label{app:ood}
In Table \ref{tab:ood} we report the score of \system{} in an out-of-domain scenario.
Specifically, for each dataset, the examples that we retrieve for the knowledge generation step do not come from the related training set contained in \kb{}.
Instead, we fetch relevant examples from the CSQA training set equipped with the ECQA knowledge explanations \cite{aggarwal-etal-2021-explanations}. This setup ensures that the examples that we provide to the LLMs do not share the same distribution as the input questions for evaluation. In the table, under every dataset, the \textit{ECQA} column reports the score of retrieving examples from the CSQA training set equipped with the explanations contained in ECQA (out-of-domain), while the \textit{ZKB} column reports the result of retrieving examples from the related training set contained in \kb{} (in-domain). From the table, we can see that the out-of-domain results are lower than the in-domain ones by an average of only 1.6 points across all the LLMs, highlighting the scalability of our approach even in scenarios in which gold or silver knowledge annotations are not available for a certain domain.

\begin{table*}[h]
    \resizebox{\textwidth}{!}{%
        \begin{tabular}{lcccccccccccccc|cc}
            \toprule
            \textbf{Model} & \multicolumn{2}{c}{\textbf{ARC-C}} & \multicolumn{2}{c}{\textbf{ARC-E}} & \multicolumn{2}{c}{\textbf{OBQA}} & \multicolumn{2}{c}{\textbf{PIQA}} & \multicolumn{2}{c}{\textbf{QASC}} & \multicolumn{2}{c}{\textbf{CSQA2}} & \multicolumn{2}{c}{\textbf{WG}} & \multicolumn{2}{c}{\textbf{AVG}}
            \\
            \cmidrule(lr){2-17}

                           & ECQA & ZKB 
                           & ECQA & ZKB 
                           & ECQA & ZKB 
                           & ECQA & ZKB 
                           & ECQA & ZKB
                           & ECQA & ZKB
                           & ECQA & ZKB
                           & ECQA & ZKB
            \\
            \midrule
            Mistral v2     & 74.4 & \textbf{75.2}
                           & 86.9 & \textbf{87.4} 
                           & 73.6 & \textbf{75.8}
                           & 77.6 & \textbf{80.2} 
                           & \textbf{68.6} & 68.3
                           & 65.6 & \textbf{67.5}
                           & 59.1 & \textbf{60.7}
                           & 72.3 & \textbf{73.6}
            \\
            Phi-3-Small    & 90.9 & \textbf{91.6}
                           & 97.2 & \textbf{97.7}
                           & 89.8 & \textbf{91.2}
                           & 87.4 & \textbf{88.1}
                           & \textbf{81.1} & 81.0
                           & 72.1 & \textbf{74.6}
                           & 78.4 & \textbf{81.0} 
                           & 85.3 & \textbf{86.4}
            \\
            Llama-3        & 82.2 & \textbf{83.5}
                           & 92.4 & \textbf{92.9}
                           & 76.8 & \textbf{79.6}
                           & 79.8 & \textbf{84.0}
                           & \textbf{80.2} & 79.1
                           & 68.2 & \textbf{69.4}
                           & 60.8 & \textbf{63.2}
                           & 77.2 & \textbf{78.8}
            \\
            Phi-3-Mini     & 86.3 & \textbf{88.0}
                           & 95.4 & \textbf{96.0} 
                           & 84.2 & \textbf{87.8}
                           & 82.0 & \textbf{84.2} 
                           & 73.7 & \textbf{73.9} 
                           & \textbf{64.7} & 64.6
                           & 69.5 & \textbf{79.2} 
                           & 79.4 & \textbf{81.9} 
            \\
            \bottomrule
        \end{tabular}%
    }
    \caption{Out-of-domain results. The \textit{ECQA} column shows the score obtained by retrieving examples from the CSQA training set, augmented with the explanations provided by ECQA (out-of-domain), while the \textit{ZKB} column represents the results of retrieving examples from the corresponding training set within \kb{} (in-domain).}
    \label{tab:ood}
\end{table*}

\subsection{Knowledge Generated with \system{}}\label{app:human}
We provide examples of the knowledge generated by the LLMs under the \system{} framework. With reference to Table \ref{tab:knowledge-examples}, we report 5 examples of knowledge generated by Llama-3 during the knowledge generation step of \system{}. Specifically, the first four examples show instances of knowledge which were deemed \textit{relevant}, \textit{factual} and \textit{helpful} by all three human annotators, while the last example shows an item of knowledge which was annotated as \textit{relevant}, \textit{factual} and \textit{harmful}. The model demonstrates its capability to generate relevant and high-quality commonsense knowledge for answering questions. However, there are instances where the model produces useful knowledge that, while generally accurate from a commonsense perspective, is not directly related to the correct choice or may harm the final prediction. Consequently, this information, despite its correctness, fails to provide the information to answer the question correctly.

\begin{table*}[h]
    \begin{tabular}{>{\centering\arraybackslash}m{0.35\linewidth}>{\centering\arraybackslash}m{0.2\linewidth}>{\centering\arraybackslash}m{0.35\linewidth}}
        \toprule
        \textbf{Question}                                                             & \textbf{Choices}                                                                                                                                                  & \textbf{Generated Knowledge}                                      \\
        \midrule
        Where in your home would you keep a ballpoint pen when not in use?            & \begin{enumerate}[A.] \itemsep-0.3em \item backpack \item bank \item \textbf{desk drawer} \item eat cake \item office desk \end{enumerate}                        & Desk drawers are used for storing office supplies.

        Pens are office supplies.

        Desk drawers are a common place to store pens.                                                                                                                                                                                                                                                                        \\
        \midrule
        Where is a likely place for an ivy plant? & \begin{enumerate}[A.] \itemsep-0.3em \item flower pot \item shelf \item windowsill \item \textbf{outside} \item sill \end{enumerate}          & 
        Ivy is a climbing plant that thrives outdoors. 
        
        Ivy often grows on walls, fences, and other surfaces. 
        
        Ivy is commonly found growing outside.                                                                                                                                                                                                                                                                             \\
        \midrule
        Where is a control room needed to prevent wide spread disaster?               & \begin{enumerate}[A.] \itemsep-0.3em \item airbase \item prison \item mill \item \textbf{nuclear plant} \item studio \end{enumerate}                              & Nuclear plant is a place where a disaster can spread widely.

        Control room is needed to prevent widespread disaster in a nuclear plant.                                                                                                                                                                                                                                             \\
        \midrule
        When a person admits his mistakes, what are they doing?                       & \begin{enumerate}[A.] \itemsep-0.3em \item accident \item learn to swim \item thank god \item feel relieved \item \textbf{act responsibly} \end{enumerate}        & Admitting mistakes means taking responsibility for one's actions.

        Taking responsibility is a sign of acting responsibly.                                                                                                                                                                                                                                                                \\
        \midrule
        \midrule
        What do the feathers look like on birds found in the rainforest?              & \begin{enumerate}[A.] \itemsep-0.3em \item pretty smart \item singing \item dark \item \textbf{very colourful} \item light \end{enumerate} & 
        Rainforests are known for their dense and dark environment. 
        
        Dark colors help birds blend in with their surroundings.                                                                                                                                                                                                                                                           \\
        \bottomrule
    \end{tabular}
    \caption{Examples of knowledge generated by Llama-3 during the knowledge generation step of \system{}. The first four examples feature explanations that have been deemed to be relevant, factual and helpful by all three human annotators; the last example features explanations that have been judged to be relevant, factual and harmful by all three human annotators. The gold answers for the questions are in \textbf{bold}.}
    \label{tab:knowledge-examples}
\end{table*}

\end{document}